%% file: _main.tex
\input{_constants}
\arxiv 

\pdfoutput=1
\documentclass[10pt,twocolumn,letterpaper]{article}
\input{cvpr_header}

\unless\ifarxiv \myexternaldocument{_supplementary} \fi

\begin{document}
\title{\paperTitle}
\author{\authorBlock}
\twocolumn[{
\renewcommand\twocolumn[1][]{#1}
\maketitle
\begin{center}
    \captionsetup{type=figure}
    \vspace{-3mm}
    \includegraphics[width=\textwidth]{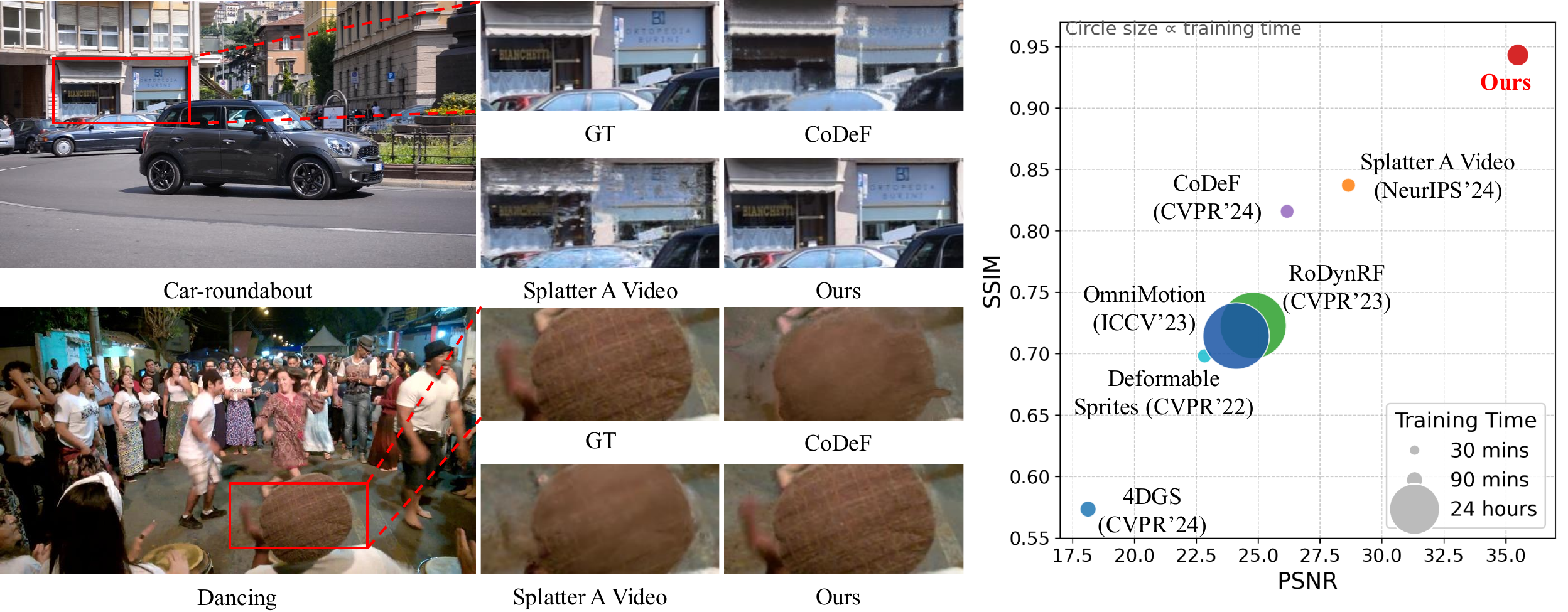}
    \vspace{-6mm}
    \captionof{figure}{
    \textbf{State-of-the-art video reconstruction quality on DAVIS~\cite{pont20172017} dataset.} Our Adaptive Gabor representation achieves superior rendering quality (PSNR: 35.49 dB, SSIM: 0.9433) while preserving fine details and temporal consistency. \emph{(Left)} Qualitative comparisons demonstrate sharper textures in challenging regions (car windows, drum surface) compared to CoDeF~\cite{ouyang2024codef} and Splatter A Video~\cite{sun2024splatter}. \emph{(Right)} Our method (\textbf{\textcolor{red}{red}} point) significantly outperforms recent baselines across all metrics, achieving 6.86 dB PSNR improvement over the second-best method with reasonable training time (circle size indicates training duration: 30 mins to 24 hours).
    }
    \label{fig:teaser}
\end{center}
}]

\input{00_abstract}
\input{01_intro}
\input{02_related}

\input{03_method}
\input{04_experiment}
\input{10_conclusion}

\newpage
\paragraph{Acknowledgements.}
This research was funded by the National Science and Technology Council, Taiwan, under Grants NSTC 112-2222-E-A49-004-MY2 and 113-2628-E-A49-023-. The authors are grateful to Google, NVIDIA, and MediaTek Inc. for their generous donations. Yu-Lun Liu acknowledges the Yushan Young Fellow Program by the MOE in Taiwan.

{\small
\bibliographystyle{ieeenat_fullname}
\bibliography{11_references}
}

\ifarxiv \clearpage \appendix \input{12_appendix} \fi

\end{document}

%% file: _constants.tex
\def\paperID{7946}
\def\confName{CVPR}
\def\confYear{2026}

\def\paperTitle{
AdaGaR: Adaptive Gabor Representation for Dynamic Scene Reconstruction}

\def\authorBlock{
    Jiewen Chan$^{1}$
    \quad
    Zhenjun Zhao$^{2}$
    \quad
     Yu-Lun Liu$^{1}$\vspace{0.5em}
    \\
    \centerline{$^1$National Yang Ming Chiao Tung University \quad $^2$University of Zaragoza}
}

\newif\ifreview 
\newif\ifarxiv \newcommand{\arxiv}{\arxivtrue}
\newif\ifcamera 
\newif\ifrebuttal 

%% file: cvpr_header.tex
\ifreview \usepackage[review]{cvpr} \fi
\ifarxiv \usepackage[pagenumbers]{cvpr} \fi
\ifrebuttal \usepackage[rebuttal]{cvpr} \fi
\ifcamera \usepackage{cvpr} \fi

\input{_macros}  

\usepackage{xr-hyper}

\makeatletter
\newcommand*{\addFileDependency}[1]{
  \typeout{(#1)}
  \@addtofilelist{#1}
  \IfFileExists{#1}{}{\typeout{No file #1.}}
}

\makeatother
\newcommand*{\myexternaldocument}[1]{
    \externaldocument{#1}
    \addFileDependency{#1.tex}
    \addFileDependency{#1.aux}
}

\definecolor{cvprblue}{rgb}{0.21,0.49,0.74}
\usepackage[pagebackref,breaklinks,colorlinks,allcolors=cvprblue]{hyperref}
\usepackage[capitalize]{cleveref}
\crefname{section}{Sec.}{Secs.}
\crefname{table}{Table}{Tables}
\crefname{figure}{Fig.}{Figs.}

\ifarxiv \crefname{appendix}{App.}{Apps.}
\else \crefname{appendix}{Suppl.}{Suppls.} \fi

%% file: _macros.tex
\usepackage{graphicx}	
\usepackage{amsmath}	
\usepackage{amssymb}	
\usepackage{booktabs}
\usepackage{times}
\usepackage{microtype}
\usepackage{epsfig}
\usepackage{caption}
\usepackage{float}
\usepackage{placeins}
\usepackage{color, colortbl}
\usepackage{stfloats}
\usepackage{enumitem}
\usepackage{tabularx}
\usepackage{xstring}
\usepackage{multirow}
\usepackage{xspace}
\usepackage{url}
\usepackage{subcaption}
\usepackage{xcolor}
\usepackage[hang,flushmargin]{footmisc}
\usepackage{makecell}

\ifcamera \usepackage[accsupp]{axessibility} \fi

% ...

\ifarxiv  \fi

\newcommand{\R}[1]{{%
    \textbf{%
        \ifstrequal{#1}{1}{\textcolor{red}{R#1}}{%
        \ifstrequal{#1}{2}{\textcolor{blue}{R#1}}{%
        \ifstrequal{#1}{3}{\textcolor{magenta}{R#1}}{%
        \ifstrequal{#1}{4}{\textcolor{teal}{R#1}}{%
                           \textcolor{cyan}{R#1}%
        }}}}%
    }%
}}

\makeatletter
\renewcommand{\paragraph}{%
    \@startsection{paragraph}{4}%
    {\z@}{-0.5em}{-0.5em}%
    {\normalfont\normalsize\bfseries}%
}
\makeatother

%% file: 00_abstract.tex
\begin{abstract}
Reconstructing dynamic 3D scenes from monocular videos requires simultaneously capturing high-frequency appearance details and temporally continuous motion. Existing methods using single Gaussian primitives are limited by their low-pass filtering nature, while standard Gabor functions introduce energy instability. Moreover, lack of temporal continuity constraints often leads to motion artifacts during interpolation.
We propose \textbf{AdaGaR}, a unified framework addressing both frequency adaptivity and temporal continuity in explicit dynamic scene modeling. We introduce Adaptive Gabor Representation, extending Gaussians through learnable frequency weights and adaptive energy compensation to balance detail capture and stability. For temporal continuity, we employ Cubic Hermite Splines with Temporal Curvature Regularization to ensure smooth motion evolution. An Adaptive Initialization mechanism combining depth estimation, point tracking, and foreground masks establishes stable point cloud distributions in early training.
Experiments on Tap-Vid DAVIS demonstrate state-of-the-art performance (PSNR 35.49, SSIM 0.9433, LPIPS 0.0723) and strong generalization across frame interpolation, depth consistency, video editing, and stereo view synthesis.
Project page: \url{https://jiewenchan.github.io/AdaGaR/}
\end{abstract}

%% file: 01_intro.tex
\vspace{-3mm}
\section{Introduction}
\label{sec:intro}

Reconstructing dynamic 3D scenes from monocular videos is a fundamental challenge in computer vision with wide applications in VR, AR, and film production. The key difficulty lies in jointly achieving temporal continuity and rich frequency representation: real-world scenes demand smooth motion over time while preserving high-frequency textures that define appearance.

Existing approaches ~\cite{qingming2025modgs,hu2024gauhuman,zhang2024bags,kocabas2024hugs,bui2025mobgs,stearns2024dynamic} fall into two camps. Gaussian-based primitives provide fast, explicit modeling but suffer from strong low-pass filtering, which suppresses high-frequency detail. Introducing frequency modulation~\cite{wurster2024gabor} (\eg Gabor-like representations) can enhance texture fidelity but often destabilizes energy balance and rendering quality. Moreover, many methods lack explicit temporal constraints, leading to motion discontinuities and geometric tearing, especially under rapid motion or occlusions.

To address these gaps, we propose {\it\textbf{AdaGaR}} (\underline{Ada}ptive \underline{G}abor \underline{R}epresentation for Dynamic Scene Reconstruction), a unified framework that jointly optimizes time and frequency in explicit dynamic representations. Our core idea is to separate and yet tightly couple two orthogonal aspects: (i) frequency adaptivity via a learnable Adaptive Gabor Representation that balances high- and low-frequency components while maintaining energy stability; and (ii) temporal continuity via Cubic Hermite Splines with Temporal Curvature Regularization, constraining motion trajectories for smooth evolution. An Adaptive Initialization further bootstraps stable, temporally coherent geometry at early training.

We validate AdaGaR on Tap-Vid~\cite{pont20172017}, achieving state-of-the-art video reconstruction and strong generalization to frame interpolation, depth consistency, video editing, and stereo view synthesis, as shown in \cref{fig:teaser}. This work provides a compact, end-to-end solution for modeling both time and frequency in explicit dynamic representations, with potential to guide future developments in frequency-aware dynamic modeling.

Our main contributions are summarized as follows:
\begin{itemize}
    \item We propose a novel \textit{Adaptive Gabor Representation} that extends traditional Gaussians to the frequency domain,~\cref{fig:motivation}, which is (i) \textit{frequency-adaptive}, (ii) \textit{energy-stable}, and (iii) \textit{capable of capturing high-frequency texture details} while automatically adjusting between high and low-frequency components according to scene requirements;
    
    \item We introduce \textit{Temporal Curvature Regularization} with \textit{Cubic Hermite Spline} interpolation, which accurately and effectively ensures geometric and motion continuity in the temporal dimension, achieving smooth temporal evolution and avoiding interpolation artifacts;
    
    \item We present an \textit{Adaptive Initialization} mechanism that combines depth estimation, point tracking, and foreground masks to establish stable and temporally consistent point cloud distributions, significantly improving training efficiency and final reconstruction quality.
    
\end{itemize}

%% file: 02_related.tex
\section{Related Work}
\label{sec:related}
\paragraph{Dynamic 3D Gaussian Splatting.}
3D Gaussian Splatting (3DGS)\cite{kerbl20233d} has inspired extensive research on dynamic scene extensions. Early work\cite{luiten2024dynamic} used time-dependent MLPs for deformation. Recent canonical space approaches~\cite{wu20244d,yang2024deformable,lu20243d,bae2024per,lin2024gaussian,huang2024sc,ho2025ted,fan2025spectromotion} employ deformation networks to handle compression and specular dynamics. Temporal modeling strategies include flow-guided methods~\cite{zhu2024motiongs}, neural features~\cite{li2024spacetime}, temporal slicing~\cite{duan20244d}, spatial-temporal regularization~\cite{li2024st,chien2025splannequin}, and hash encoding~\cite{xu2024grid4d}. Specialized applications target autonomous driving~\cite{zhou2024drivinggaussian,yan2024street,shen2024driveenv}, sparse reconstruction~\cite{mihajlovic2024splatfields}, unconstrained capture~\cite{kim20244d,shih2025prior}, acceleration~\cite{tu2025speedy,zhan2025cat}, and motion blur~\cite{wu2024deblur4dgs}. Most similar to our work, SplineGS~\cite{park2025splinegs} applies Cubic Hermite splines in multi-view settings. In contrast, we combine \emph{Cubic Hermite splines} with \emph{Gabor-based primitives} for \emph{monocular} videos without camera pose estimation, introducing Temporal Curvature Regularization for physically plausible motion.
\paragraph{Frequency-Adaptive Rendering.}
Traditional 3D Gaussian kernels act as low-pass filters, limiting high-frequency detail representation. Anti-aliasing methods for 3DGS include multi-scale filtering~\cite{yu2024mip,yan2024multi}, analytical integration~\cite{liang2024analytic}, and opacity field derivation~\cite{yu2024gaussian}. NeRF frequency-aware approaches employ cone-tracing~\cite{barron2023zip}, frequency regularization~\cite{xie2024g,yang2023freenerf}, frequency decomposition~\cite{he2024freditor}, and structure-noise separation~\cite{qu2024lush}. Gabor representations in neural rendering~\cite{almughrabi2024momentsnerf,watanabe20253d,wurster2024gabor} build on procedural graphics foundations~\cite{lagae2009procedural,galerne2012gabor}. Alternative primitives include exponential functions~\cite{hamdi2024ges}, surfels~\cite{huang20242d}, and Beta kernels~\cite{liu2025deformable}. However, existing Gabor approaches target \emph{static} scenes with \emph{fixed} frequencies. Our Adaptive Gabor Representation extends to \emph{dynamic} videos with \emph{learnable} frequency weights and graceful degradation to standard Gaussians.
\paragraph{Temporal Modeling and Spline Representations.}
Classical splines~\cite{farin2002curves,liu2014skinning} provide smooth temporal interpolation. Recent neural rendering incorporates splines through Hermite formulations~\cite{chugunov2024neural}, B-splines~\cite{wang2024vidu4d,park2025splinegs}, and time-modulated weights~\cite{grega2024neural}. Coarse-fine decomposition methods~\cite{bae2024per,yang2023real,yan2024ds} separate temporal scales. Flow-guided approaches~\cite{liang2024flowvid,zhou2024upscale,zhu2024motiongs,ma2024trailblazer} leverage optical flow constraints. Alternative temporal models include Kalman filtering~\cite{zhan2024kfd}, neural trajectories~\cite{kratimenos2024dynmf}, frame interpolation~\cite{reda2022film,huang2022real}, and robust dynamic fields~\cite{liu2023robust}. Unlike implicit smoothness from architecture or training, we \emph{explicitly} enforce smoothness through Temporal Curvature Regularization based on second-order derivatives, ensuring physically plausible motion with geometric interpretability.
\paragraph{Video Representations and Canonical Spaces.}
Canonical space methods enable temporally consistent processing through layered atlases~\cite{kasten2021layered}, deformation fields~\cite{ouyang2024codef,chen2024narcan}, and canonical volumes~\cite{wang2023tracking}. Implicit neural video representations~\cite{chen2021nerv,chen2023hnerv,li2022nerv,kim2024snerv,shin2024efficient,saethre2024combining,kwan2024nvrc} achieve compression through image-wise functions. Explicit representations employ 4D Gaussians~\cite{wu20244d}, 2D feature streams~\cite{wang2024videorf}, layer decomposition~\cite{shrivastava2024video}, learned quantization~\cite{li2024neural}, hash encoding~\cite{chen2025dash}, and scene inpainting~\cite{wu2025aurafusion360}. Video Gaussian Splatting methods target monocular~\cite{qingming2025modgs,hu2024gauhuman,zhang2024bags,kocabas2024hugs,bui2025mobgs,stearns2024dynamic,lin2025longsplat,hou20253d} and multi-view~\cite{chen2024mvsplat,liu2024mvsgaussian} settings. Our approach operates in an \emph{orthographic camera coordinate system}~\cite{sun2024splatter}, eliminating pose estimation while maintaining explicit 3D structure through Gabor primitives for high-frequency preservation and versatile applications.
\paragraph{Monocular Depth and Motion Estimation.}
Foundation models provide robust monocular priors. Depth estimation methods~\cite{yang2024depth,ke2024repurposing,bochkovskii2024depth,hu2024metric3d,piccinelli2024unidepth,li2024patchfusion,ranftl2021vision,wang2024depth} achieve zero-shot generalization through synthetic training, diffusion repurposing, and multi-dataset learning. Point tracking methods~\cite{karaev2025cotracker3,doersch2024bootstap,cho2024local,doersch2023tapir,karaev2024cotracker,wang2024sea,harley2025alltracker} enable dense correspondence through pseudo-labeling, self-supervision, and local correlation. Unlike prior work using these signals independently, our \emph{adaptive initialization} jointly reasons about depth, motion, and segmentation for geometrically and temporally consistent initialization.

%% file: 03_method.tex
\begin{figure}[t]
    \centering
    \includegraphics[width=\linewidth]{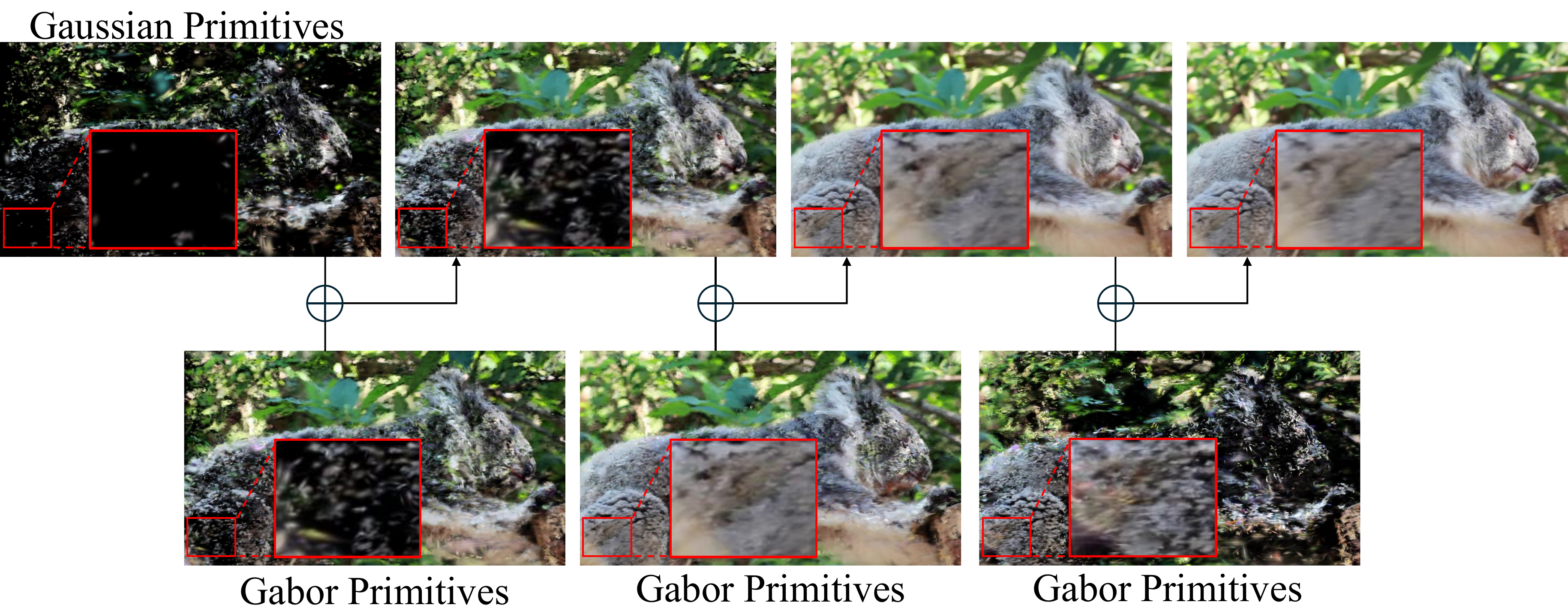}
    \vspace{-6mm}
    \caption{
    \textbf{Hierarchical frequency adaptation.} 
    Our primitives adaptively transition from Gaussian (topleft) to Gabor (bottom), enabling coarse-to-fine reconstruction. Each primitive learns its optimal frequency response via learnable weights $\omega_i$, achieving both geometric stability and texture detail in a unified framework.
    }
    \label{fig:motivation}
\end{figure}

\section{Preliminary: 3D Gaussian Splatting}
\label{sec:preliminary}

3D Gaussian Splatting (3DGS)~\cite{kerbl20233d} represents a 3D scene as a collection of parameterized Gaussian primitives $\{\mathcal{G}_k \mid k=1,\ldots,N\}$. Each $\mathcal{G}_k$ has center $\boldsymbol{\mu}_k \in \mathbb{R}^3$, covariance $\boldsymbol{\Sigma}_k \in \mathbb{R}^{3\times3}$, opacity $\alpha_k \in [0,1]$, and color $\mathbf{c}_k$. The density is
\[
\mathcal{G}_k(\mathbf{x}) = \exp\left(-\tfrac{1}{2}(\mathbf{x}-\boldsymbol{\mu}_k)^\top \boldsymbol{\Sigma}_k^{-1}(\mathbf{x}-\boldsymbol{\mu}_k)\right),
\]
with $\boldsymbol{\Sigma}_k = \mathbf{R}_k\mathbf{S}_k\mathbf{S}_k^\top\mathbf{R}_k^\top$.

Rendering projects Gaussians onto the image plane and accumulates color via front-to-back blending:
\[
C(\mathbf{x}) = \sum_{k=1}^{K} T_k \alpha_k \mathbf{c}_k,\quad T_k = \prod_{j<k}(1-\alpha_j).
\]

A key limitation is that a single Gaussian acts as a low-pass filter, constraining high-frequency textured detail. To address this, we introduce Gabor kernels as periodic extensions of Gaussians to enhance spatial frequency representation.

\section{Method}
\label{sec:method}
\subsection{Overview}
\begin{figure*}[t]
    \centering
    \vspace{-3mm}
    \includegraphics[width=\textwidth]{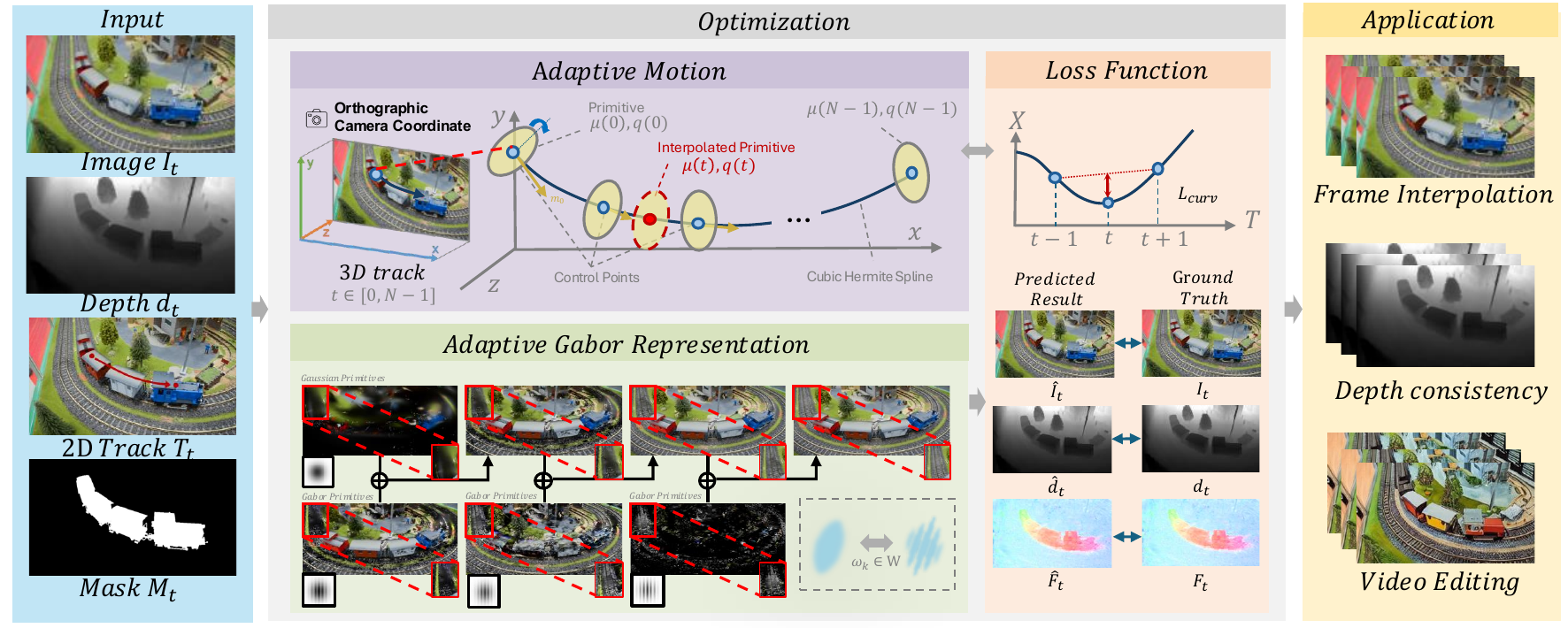}
    \vspace{-6mm}
    \caption{
    \textbf{Method overview.} Our approach represents dynamic videos as Adaptive Gabor primitives with temporally smooth motion. (\emph{Input}) Multi-modal supervision from RGB, depth, tracking, and masks. (\emph{Optimization}) Two core components: (1) \emph{Adaptive Motion}: Cubic Hermite splines model primitive trajectories with control points {$\mu(t)$, $q(t)$} in orthographic camera space, ensuring C$^1$ continuity. (2) \emph{Adaptive Gabor Representation}: Learnable frequency weights $\omega_k$ enable primitives to adaptively span from Gaussian (low-freq) to Gabor (high-freq), achieving hierarchical detail reconstruction. (\emph{Loss}) Joint optimization via RGB, depth, flow supervision, and curvature regularization $L_{curv}$. (\emph{Application}) Supports frame interpolation, depth consistency, and video editing.
    }
    \label{fig:pipeline}
\end{figure*}

We present AdaGaR, an explicit 3D video representation that preserves high-frequency appearance while ensuring temporally smooth motion. As illustrated in~\cref{fig:pipeline}, The video is modeled as a set of dynamic Adaptive Gabor primitives in an orthographic camera coordinate system, where spatial texture and structure are encoded by the primitives and temporal evolution is interpolated with Cubic Hermite Splines to guarantee geometric and temporal consistency.
Adaptive Gabor Representation extends Gaussian primitives with learnable frequency weights and energy compensation, enabling frequency-adaptive detail capture while maintaining energy stability. Coupled with temporal curvature regularization and multi-supervision losses, our approach delivers high visual quality and robust temporal consistency, with strong applicability to frame interpolation, depth consistency, video editing, and related tasks.

\subsection{Adaptive Gabor Video Representation}

\paragraph{Camera Coordinate Space.}
Inspired by~\cite{wang2023tracking} and~\cite{sun2024splatter}, we adopt an orthographic camera coordinate system that maps width, height, and depth to the $X$, $Y$, and $Z$ axes, enabling a direct orthogonal representation of the 3D video structure. This avoids costly camera pose estimation and motion disentangling, treating camera motion and object motion as a single type of dynamic variation. The video is represented as a collection of dynamic adaptive Gabor primitives, each encoding spatial position, temporal variation, and frequency response, rendered from a fixed identity pose. 

\paragraph{Adaptive Gabor Representation.}
\label{sec:adaptive_gabor_representation}
To introduce high-frequency details on the image plane, the Gabor function can be viewed as a periodic extension of the Gaussian function. Its general 2D form can be defined as:
\begin{equation}
\mathcal{G}_{\text{Gabor}}(\mathbf{x}) = \exp\left(-\frac{1}{2}||\mathbf{x}-\boldsymbol{\mu}||^2_{\boldsymbol{\Sigma}^{-1}}\right) \cos(\mathbf{f}^\top\mathbf{x} + \phi),
\end{equation}
where $\mathbf{x} = (x, y)^\top$ denotes the image plane coordinates, $\mathbf{f} = (f_x, f_y)^\top$ is the center frequency vector, and $\phi$ represents the phase offset. This structure introduces a sinusoidal modulation within the Gaussian envelope, enabling the distribution to simultaneously capture local directional textures and high-frequency detail variations.

To model richer frequency components, multiple Gabor waves can be combined into a weighted superposition:
\begin{equation}
S(\mathbf{x}) = \sum_{i=1}^{N} \omega_i \cos(f_i \langle \mathbf{d}_i, \mathbf{x} \rangle + \phi_i),
\end{equation}
where $\omega_i \in \mathbb{R}$ denotes the amplitude weight, $f_i \in \mathbb{R}^+$ represents the frequency magnitude, and $\mathbf{d}_i \in \mathbb{R}^2$ with $||\mathbf{d}_i||_2 = 1$ is the frequency direction unit vector. This structure generates spatially periodic texture variations, which produce richer textural details when combined with Gaussians.

While the Gabor structure enhances detail representation, fixed-amplitude cosine modulation disrupts the energy stability of the Gaussian. To address this, we propose \textit{Adaptive Gabor}, which automatically adjusts the intensity based on the wave energy and naturally degrades to a Gaussian in extreme cases. We extend the original opacity expression to $\alpha_{\text{Gabor}} = \mathcal{G}(\mathbf{x}) \cdot S(\mathbf{x})$. In practice, we set the phase terms $\phi_i = 0$, yielding:
\begin{equation}
S(\mathbf{x}) = \sum_{i=1}^{N} \omega_i \cos(f_i \langle \mathbf{d}_i, \mathbf{x} \rangle),
\end{equation}
where we fix the frequency parameters $f_i \in \{1, 2\}$, corresponding to two orthogonal base frequency waveforms. The amplitude weights $\omega_i \in [0, 1]$ are the introduced learnable parameters for the Gabor structure, adjusting the energy weights of different frequency components. The direction unit vectors $\mathbf{d}_i$ are shared with the spatial orientation of the original Gaussian, ensuring consistency between frequency modulation and Gaussian shape orientation.

To prevent overall intensity attenuation when $\sum_i \omega_i < 1$, we introduce a compensation term $b$:
\begin{equation}
S_{adap}(\mathbf{x}) = b + \frac{1}{N} \sum_{i=1}^{N} \omega_i \cos(f_i \langle \mathbf{d}_i, \mathbf{x} \rangle),
\end{equation}
\begin{equation} \label{eq:b_equation}
b = \gamma + (1 - \gamma)\left(1 - \frac{1}{N}\sum_{i=1}^{N} \omega_i\right),
\end{equation}
where $\gamma \in [0, 1]$ is a fixed hyperparameter controlling the degradation smoothness, and the factor $1/N$ normalizes the weighted average of multiple waves to a stable range. When $\omega_i \to 0$, we have $b \to 1$, and the formulation naturally degrades to a traditional Gaussian.


\begin{figure*}[t]
\centering
    \vspace{-3mm}
\begin{minipage}[t]{0.58\textwidth}
    \centering
    \includegraphics[width=\textwidth]{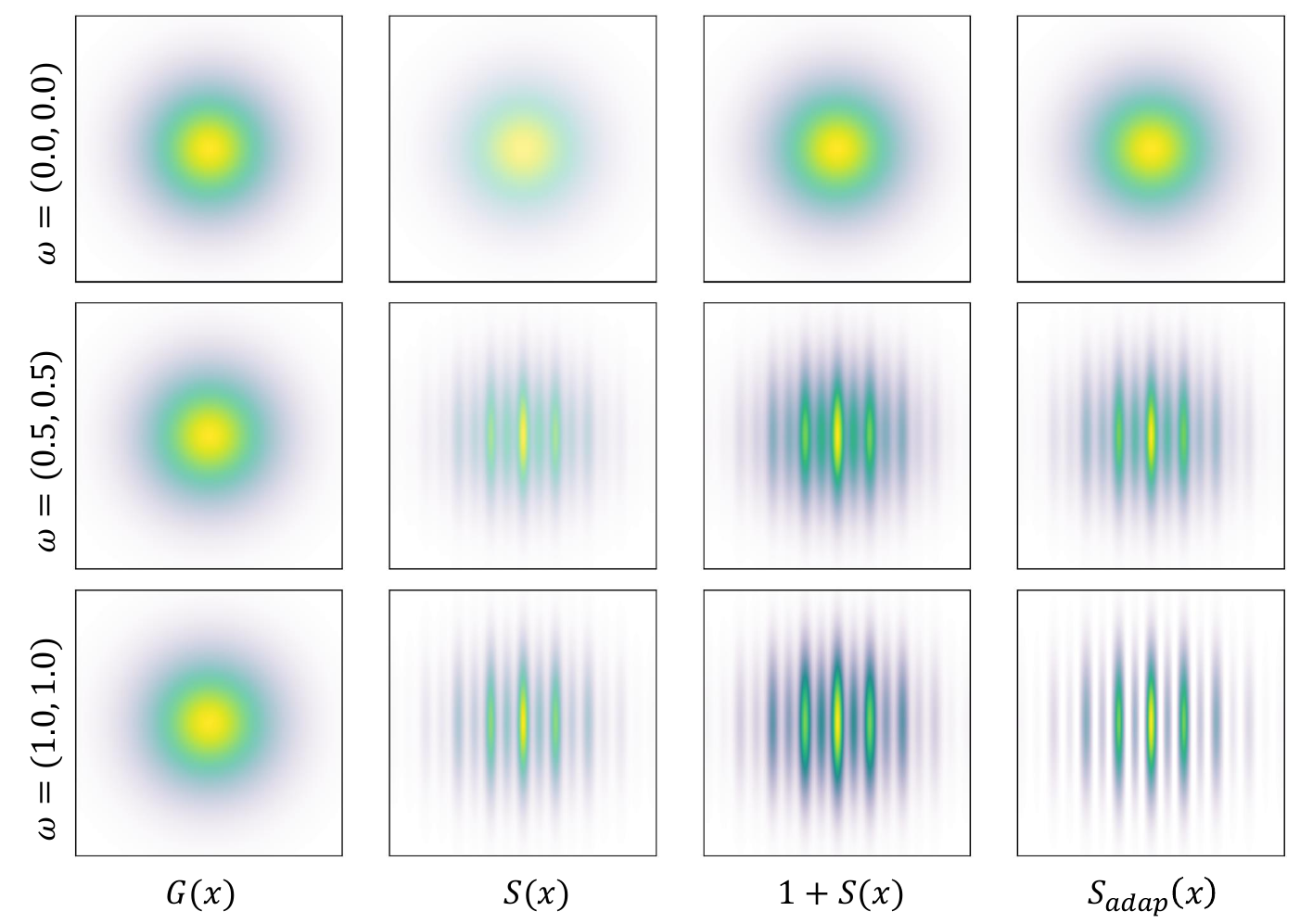}
    
    (a) $G(x)$,Gaussian, $S(x)$, sinusoid part
\end{minipage}
\hfill
\begin{minipage}[t]{0.38\textwidth}
    \centering
\vspace{-71mm}
    \includegraphics[width=\textwidth]{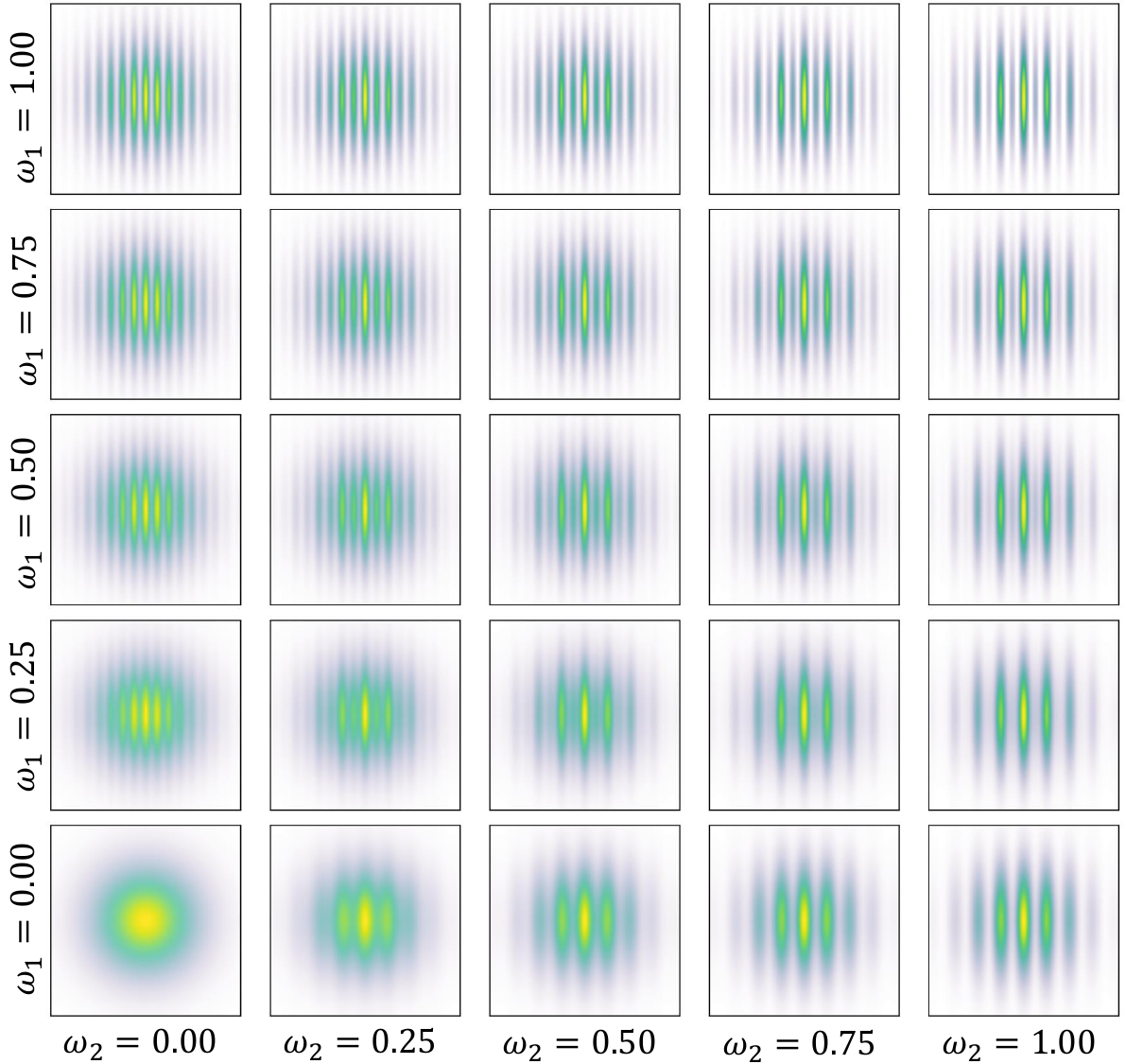}
    (b) The effect of different combinations of Gabor wave coefficients ($\omega_{0}$ and $\omega_{1}$) on spatial frequency textures.
\end{minipage}
\vspace{-2mm}
\caption{
\textbf{Adaptive Gabor formulation.} \textbf{(a) Smooth transition between Gaussian and Gabor kernels.} Our method (rightmost column, $S_\text{ours}(x)$) uses a compensation term $b$ to maintain energy stability while transitioning from pure Gaussian ($\omega=0$, top) to frequency-modulated Gabor ($\omega=1$, bottom). Naive combination $1+S(x)$ (third column) suffers from intensity artifacts. \textbf{(b) Frequency weight combinations.} Different ($\omega_0$, $\omega_1$) pairs generate diverse spatial patterns, from smooth (low $\omega$) to high-frequency textures (high $\omega$), enabling adaptive detail capture in different scene regions.
}
\label{fig:gabor_compare}
\end{figure*}



\subsection{Temporally Dynamic Adaptive Gabor}

\paragraph{Cubic Hermite Spline Interpolation.}
We use Cubic Hermite Splines~\cite{hintzen2010improved,chand2012cubic} to interpolate the temporal evolution of dynamic primitives. Given $M$ temporal keyframes at times $\{t_0, t_1, \ldots, t_{M-1}\}$ with corresponding control point positions $\{\mathbf{y}_0, \mathbf{y}_1, \ldots, \mathbf{y}_{M-1}\} \subset \mathbb{R}^3$, we define the time interval between adjacent keyframes as $\Delta_k = t_{k+1} - t_k$, and the slope as $\delta_k = (\mathbf{y}_{k+1} - \mathbf{y}_k) / \Delta_k$.
To avoid unnecessary oscillations between keyframes, we introduce an auto-slope mechanism with a monotone gate:
\begin{equation} \label{eq:ours_spline}
\mathbf{m}_k = 
\begin{cases}
\beta \cdot \frac{\delta_{k-1} + \delta_k}{2}, & \text{if } \mathrm{sign}(\delta_{k-1}) = \mathrm{sign}(\delta_k), \\
\mathbf{0}, & \text{otherwise},
\end{cases}
\end{equation}
where $\beta \in (0, 1]$ is a smoothness coefficient controlling the flatness of the interpolation curve. This design prevents reverse oscillations at keyframes and ensures visually stable interpolation.

The Hermite basis functions are defined as:
\begin{equation}
\begin{aligned}
&H_{00}(s) = 2s^3 - 3s^2 + 1, \quad H_{10}(s) = s^3 - 2s^2 + s, \\
&H_{01}(s) = -2s^3 + 3s^2, \quad\;\;\, H_{11}(s) = s^3 - s^2,
\end{aligned}
\end{equation}
where $s = (t - t_k) / \Delta_k \in [0, 1]$ is the normalized time within the interval $[t_k, t_{k+1}]$. The interpolated displacement at time $t$ is:
\begin{equation}
\begin{aligned}
\boldsymbol{\Delta}(t) = \,&H_{00}(s)\mathbf{y}_k + H_{10}(s)\Delta_k \mathbf{m}_k \\
&+ H_{01}(s)\mathbf{y}_{k+1} + H_{11}(s)\Delta_k \mathbf{m}_{k+1}.
\end{aligned}
\end{equation}

To ensure consistent geometric continuity, the final position is obtained by adding the interpolated displacement to a base position:
\begin{equation}
\boldsymbol{\mu}(t) = \boldsymbol{\mu}_{\text{base}} + \boldsymbol{\Delta}(t).
\end{equation}

\noindent\textbf{Rotation Interpolation.} We extend the same principle to temporal interpolation of rotations. For rotation parameters, we first interpolate in the $\mathfrak{so}(3)$ Lie algebra space, then convert to unit quaternions via the exponential map:
\begin{equation}
\mathbf{q}(t) = \mathrm{normalize}\big(\mathrm{normalize}(\mathbf{q}_{\text{base}}) \otimes \exp(\boldsymbol{\Delta}_{\mathbf{q}}(t))\big),
\end{equation}
where $\otimes$ denotes quaternion multiplication, and angle wrapping ensures rotation angles remain within $(-\pi, \pi]$.


\paragraph{Temporal Curvature Regularization.}
To enforce smooth temporal evolution, we introduce a curvature penalty on the trajectory at each keyframe. For non-uniform keyframes, the second-order derivative is estimated as
\begin{equation}
\mathbf{y}_k'' = \frac{2(\mathbf{d}_k^+ - \mathbf{d}_k^-)}{h_{k-1} + h_k},
\end{equation}
with $h_{k-1}=t_k-t_{k-1}$, $h_k=t_{k+1}-t_k$, $\mathbf{d}_k^+ = (\mathbf{y}_{k+1}-\mathbf{y}_k)/h_k$, $\mathbf{d}_k^- = (\mathbf{y}_k-\mathbf{y}_{k-1})/h_{k-1}$, and $D=3$. The curvature loss is
\begin{equation}
\mathcal{L}_{\text{curve}} = \frac{\sum_{k=1}^{M-2} w_k \|\mathbf{y}_k''\|_2^2}{\sum_{k=1}^{M-2} w_k D + \varepsilon},
\end{equation}
where $w_k = \tfrac{1}{2}(h_{k-1}+h_k)$ and $\varepsilon>0$ is a small constant. This term enforces smoothness by penalizing the second-order energy along time.

\subsection{Optimization}

To maintain both realistic appearance and temporal stability in dynamic scenes, we employ a multi-objective loss function that constrains appearance fidelity, motion consistency, depth geometry, and temporal smoothness.

\paragraph{Rendering Reconstruction Loss.} 
We combine $\mathcal{L}_1$ and SSIM to preserve both pixel-level accuracy and structural features:
\begin{equation}
\mathcal{L}_{\text{rgb}}(I_t, \hat{I}_t) = (1 - \lambda_{\text{ssim}})\mathcal{L}_1^{\text{rgb}}(I_t, \hat{I}_t) + \lambda_{\text{ssim}}\mathcal{L}_{\text{ssim}}^{\text{rgb}}(I_t, \hat{I}_t),
\end{equation}
where $I_t$ and $\hat{I}_t$ denote the ground-truth and predicted images at frame $t$, respectively.

\paragraph{Optical Flow Consistency Loss.} We leverage CoTracker~\cite{karaev2024cotracker} to provide cross-frame supervision. The projected positions of Adaptive Gabor primitives are aligned with 2D trajectories using a visibility-weighted $\mathcal{L}_1$ loss:
\begin{equation}
\mathcal{L}_{\text{flow}}(\hat{F}_{t_1,t_2}, F_{t_1,t_2}) = \frac{\sum_j w_j \left\| \hat{\mathbf{x}}_{t_2}^j - \mathbf{x}_{t_2}^j \right\|_1}{\sum_j w_j + \varepsilon},
\end{equation}
where $\mathbf{x}_{t_2}^j$ and $\hat{\mathbf{x}}_{t_2}^j$ are the ground-truth and predicted pixel positions of the $j$-th tracked point at frame $t_2$, and $w_j$ denotes its visibility weight. 

\paragraph{Depth Loss.} We use monocular depth estimates from DPT~\cite{ranftl2021vision} as geometric priors with scale- and shift-invariant alignment:
\begin{equation}
\mathcal{L}_{\text{depth}}(D_t, \hat{D}_t) = \left\| \gamma(D_t) - \gamma(\hat{D}_t) \right\|_1,
\end{equation}
where $\gamma(D_t) = (D_t - c_t(D_t)) / \|D_t - c_t(D_t)\|_1$ with $c_t(D_t) = \text{median}(D_t)$.

\paragraph{Total Loss.} The overall optimization objective combines all components:
\begin{equation}
\mathcal{L}_{\text{total}} = \lambda_{\text{rgb}}\mathcal{L}_{\text{rgb}} + \lambda_{\text{flow}}\mathcal{L}_{\text{flow}} + \lambda_{\text{depth}}\mathcal{L}_{\text{depth}} + \lambda_{\text{curv}}\mathcal{L}_{\text{curv}}.
\end{equation}
This multi-faceted supervision enables AdaGaR to achieve both high-fidelity rendering and temporally stable dynamic scene representation.

\subsection{Adaptive Initialization}

We propose an adaptive initialization to initialize a temporally coherent 3D point distribution early in training. It fuses multi-modal cues to generate a dense, dynamic initial point cloud from the input video, forming the geometric basis for subsequent explicit representations.
Unlike random sampling or single-frame methods, our approach adaptively adjusts sampling density according to scene motion and depth distribution, ensuring balanced foreground/background coverage.

\paragraph{Temporal–Spatial Adaptive Sampling.} For each candidate point $\mathbf{p}_i$, the sampling probability is
\begin{equation}
\label{eq:adap_init_1}
\Pi(\mathbf{p}_i) \propto \frac{1}{\tau_i+\epsilon} + \lambda_\tau \frac{1}{\rho_i+\epsilon},
\end{equation}
where $\tau_i$ is the temporal support, $\rho_i$ the local density, $\lambda_\tau \in [0,1]$ balances temporal stability and spatial uniformity, and $\epsilon>0$.

\paragraph{Grid-Based Uniform Coverage.} To ensure global coverage, we partition the image into a fixed grid $\mathcal{G}=\{G_{u,v}\}$ and modulate per-cell sampling by
\begin{equation}
\label{eq:adap_init_2}
\Pi'(\mathbf{p}_i \mid G_{u,v}) = \frac{\Pi(\mathbf{p}_i)}{1+\lambda_g C_{u,v}},
\end{equation}
with $C_{u,v}$ the cell’s cumulative samples and $\lambda_g>0$.

\paragraph{Boundary-Aware Compensation.} We further adjust for motion boundaries via
\begin{equation}
\label{eq:adap_init_3}
\Pi''(\mathbf{p}_i) = \Pi'(\mathbf{p}_i \mid G_{u,v}) \left(1+\lambda_b \|\nabla M_t(\mathbf{p}_i)\|\right),
\end{equation}
where $M_t$ is the foreground mask and $\lambda_b>0$.

This scheme yields a dense, temporally coherent initial point cloud and reduces early-stage flickering.


%% file: 04_experiment.tex
\section{Experiment}
\label{sec:method}

\subsection{Evaluation}

\paragraph{Dataset and Metrics.}
We evaluate on Tap-Vid DAVIS~\cite{pont20172017}, featuring diverse dynamic scenes and occlusions. Quantitative metrics include PSNR, SSIM, and LPIPS to assess pixel accuracy, structural fidelity, and perceptual quality.

\paragraph{Implementation Details.}
The training consists of two stages: a 500-iteration warm-up and 10K iterations for main optimization, with control points updated every 100 iterations. Experiments run on an NVIDIA RTX 4090, ~90 minutes per video sequence.

\input{tables/reconstruction}

\begin{figure*}[t]
    \centering
    \vspace{-3mm}
    \includegraphics[width=\textwidth]{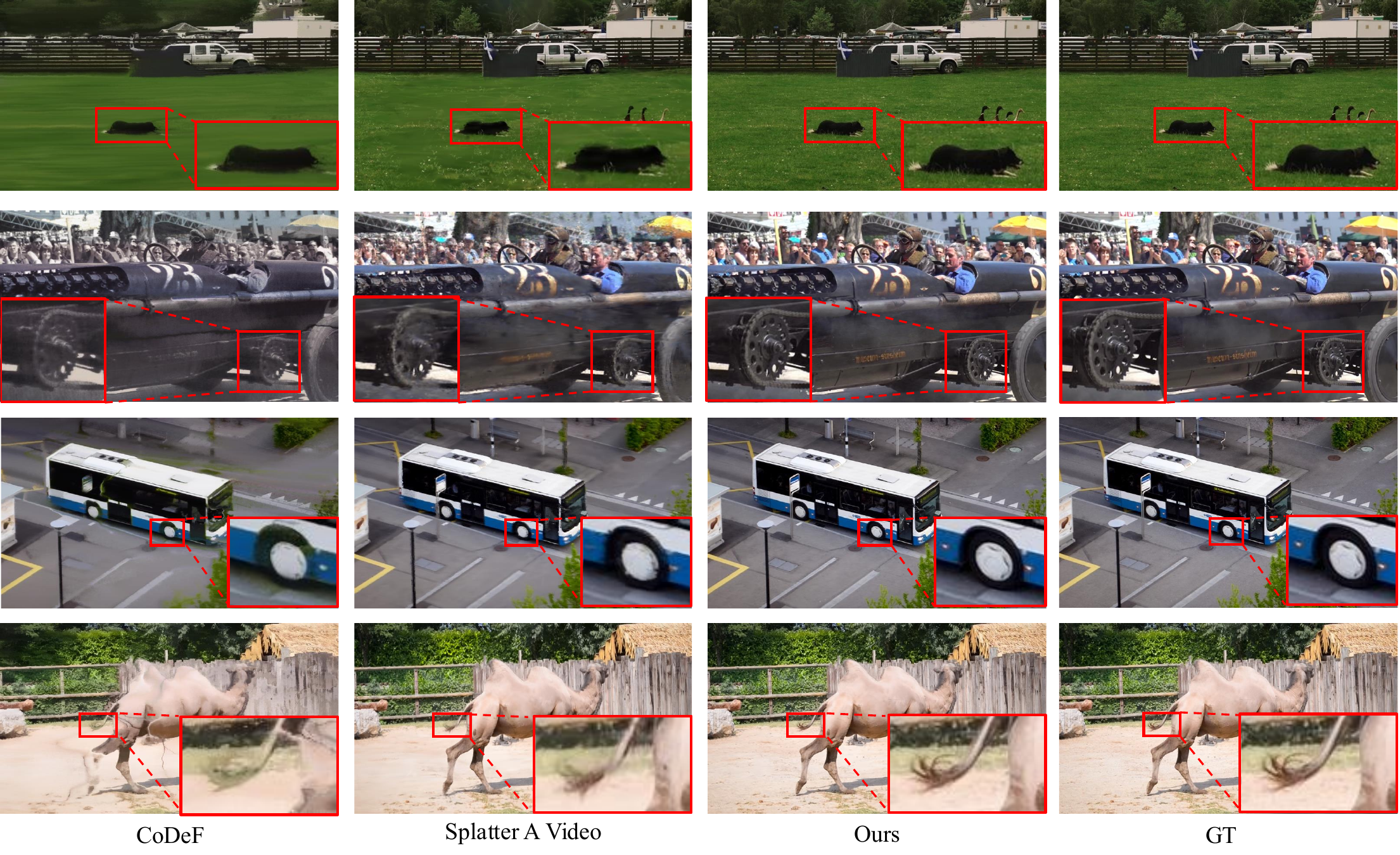}
    \vspace{-6mm}
    \caption{
    \textbf{Visual comparison on DAVIS dataset.} 
    Our method preserves finer details (fur, vehicle edges, wheel structures) and sharper motion boundaries compared to CoDeF~\cite{ouyang2024codef} and Splatter A Video~\cite{sun2024splatter}. \textcolor{red}{Red} boxes highlight key regions demonstrating our superior texture reconstruction and temporal consistency. Best viewed zoomed in.
    }
    \label{fig:visual_compare}
\end{figure*}

\paragraph{Video Reconstruction.} 
As shown in~\cref{tab:reconstruction}, our method outperforms the baselines across PSNR/SSIM/LPIPS on Tap-Vid DAVIS~\cite{pont20172017}. Compared with MLP-based representations, ours yields sharper textures and coherent motion in~\cref{fig:visual_compare}.

\begin{figure}[t]
    \centering
    \includegraphics[width=\linewidth]{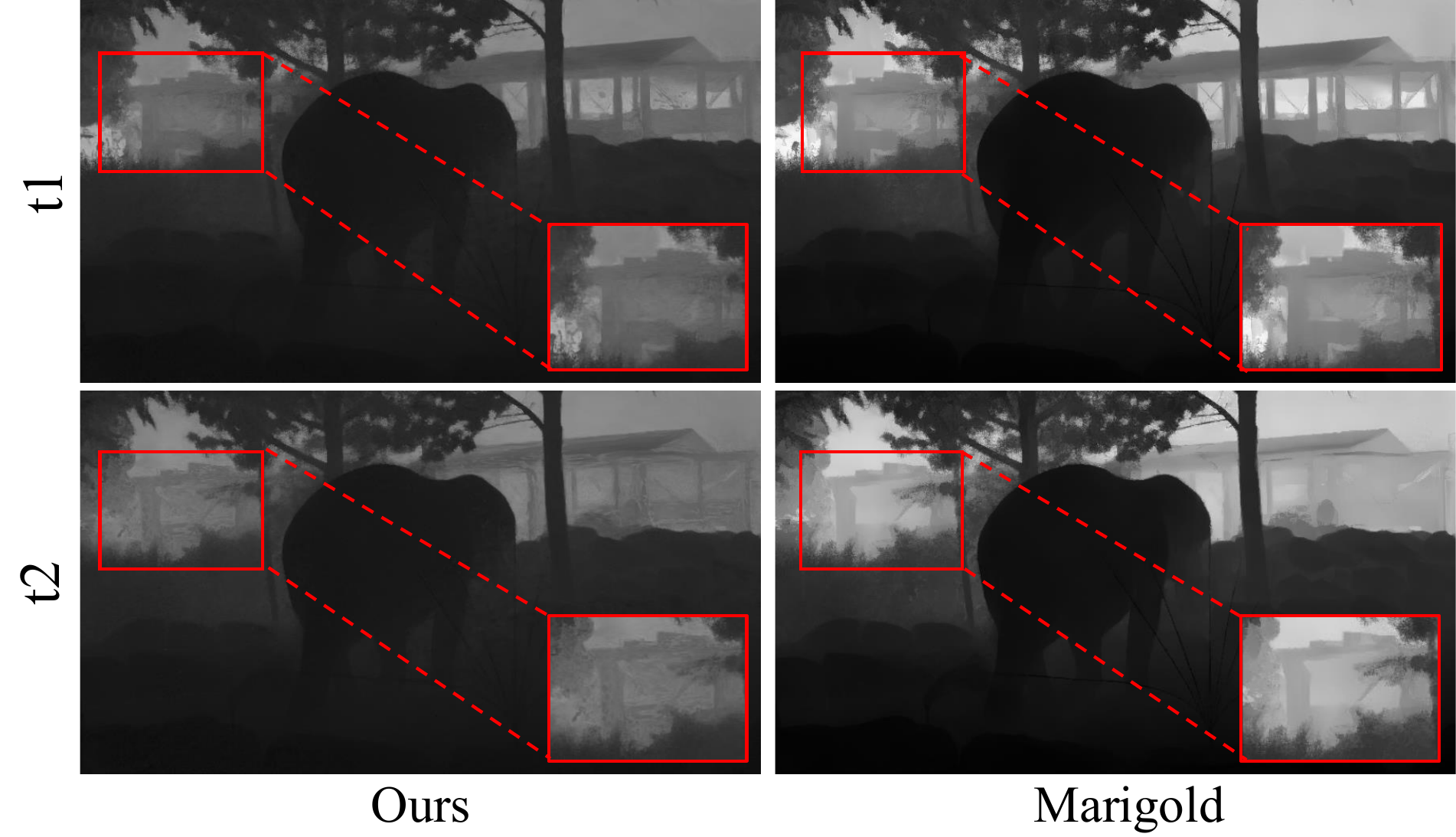}
    \vspace{-6mm}
    \caption{
    \textbf{Depth consistency across time.} \emph{(Left)} Our 3D primitive representation maintains consistent depth for static elements across frames. \emph{(Right)} While per-frame estimation (Marigold~\cite{ke2024repurposing}) shows temporal flickering (\textcolor{red}{red} boxes). Explicit 3D geometry with smooth motion modeling ensures temporal coherence essential for depth-based video applications.
    }
    \label{fig:depth_consistency}
\end{figure}

\subsection{Applications}
\paragraph{Depth Consistency.} 
We achieve stable depth distributions over time, substantially reducing depth flicker and boundary misalignment, and outperforming per-frame optimizers, as shown in~\cref{fig:depth_consistency}.

\begin{figure*}[t]
    \centering
    \vspace{-3mm}
    \includegraphics[width=\textwidth]{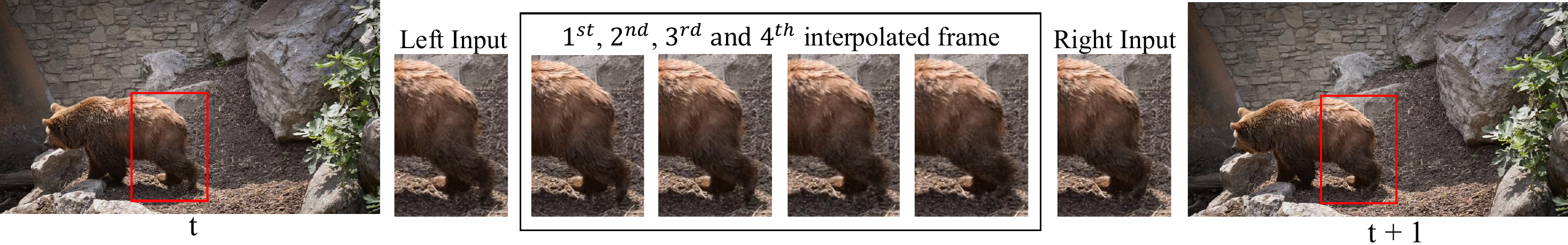}
    \vspace{-6mm}
    \caption{
    \textbf{Frame interpolation results.} Our method generates temporally smooth intermediate frames between input keyframes $t$ and $t+1$ by querying Cubic Hermite splines at fractional timestamps. The interpolated sequence ($\text{1}^\text{st}$ through $\text{4}^\text{th}$ frames) maintains consistent fur texture details and natural motion without ghosting artifacts. Red boxes show the preservation of high-frequency details throughout the interpolation. This demonstrates our method's ability to produce continuous motion with $C^1$ smoothness via curvature-regularized spline trajectories. Please refer to the supplementary video for full temporal coherence.
    }
    \label{fig:interpolation}
\end{figure*}

\paragraph{Frame Interpolation.} 
We generate smooth intermediate frames between keyframes using cubic Hermite splines with curvature regularization, preserving texture detail and avoiding boundary artifacts, as shown in~\cref{fig:interpolation}.

\begin{figure}[t]
    \centering
    \includegraphics[width=\linewidth]{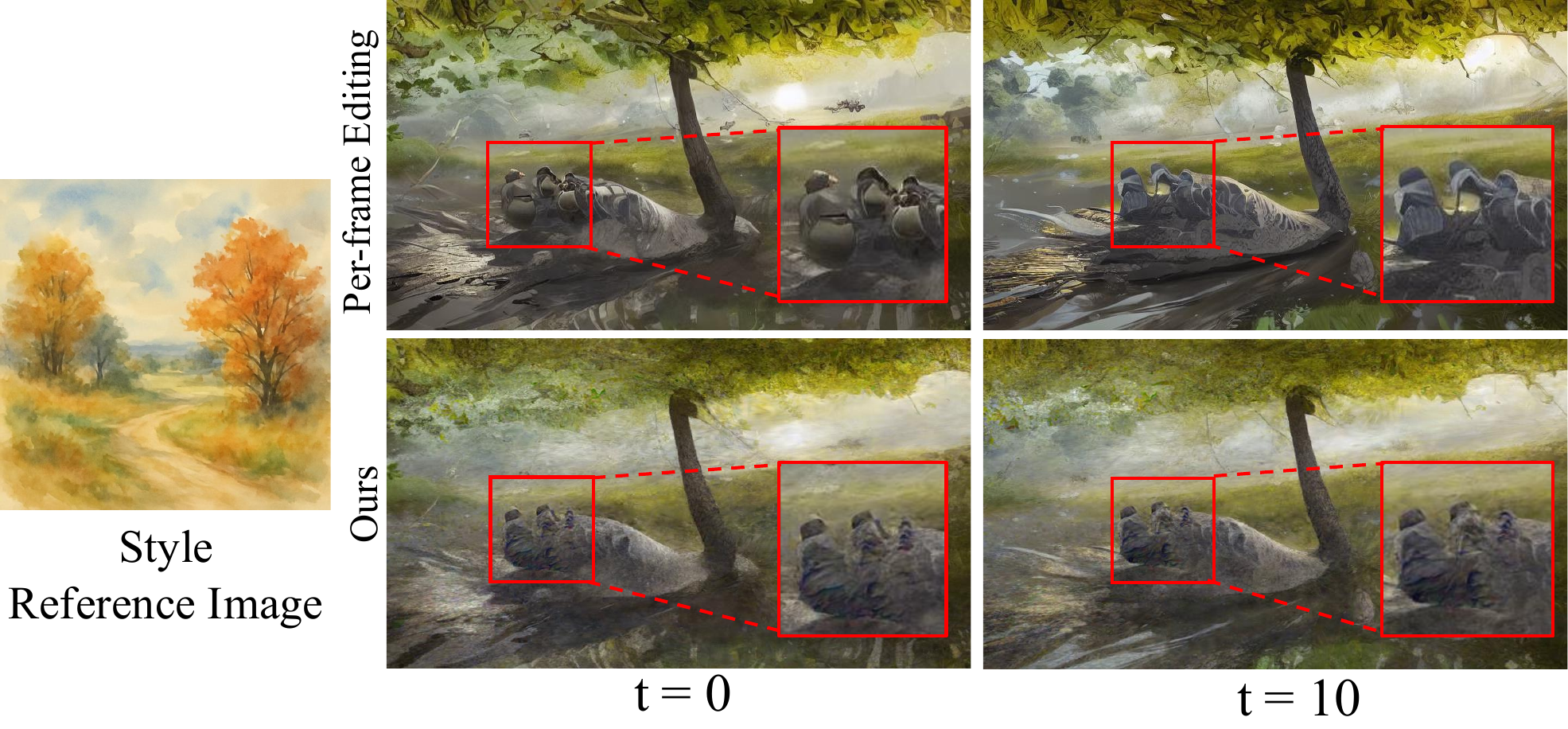}
    \vspace{-6mm}
    \caption{
    \textbf{Temporally consistent video editing.} 
    (\emph{Top}) Per-frame editing causes temporal flickering with inconsistent styles between frames. (\emph{Bottom}) Our canonical space editing maintains temporal consistency by applying style transfer to shared Adaptive Gabor primitives, ensuring identical treatment of scene elements across time while preserving motion dynamics. \textcolor{red}{Red} boxes highlight key differences. Please see the supplementary video.
    }
    \label{fig:video_editing}
\end{figure}

\paragraph{Video Editing.} 
In canonical space, style transfers remain temporally coherent by acting on shared Adaptive Gabor primitives, reducing style drift and flicker, as shown in~\cref{fig:video_editing}.

\begin{figure}[t]
    \centering
    \includegraphics[width=\linewidth]{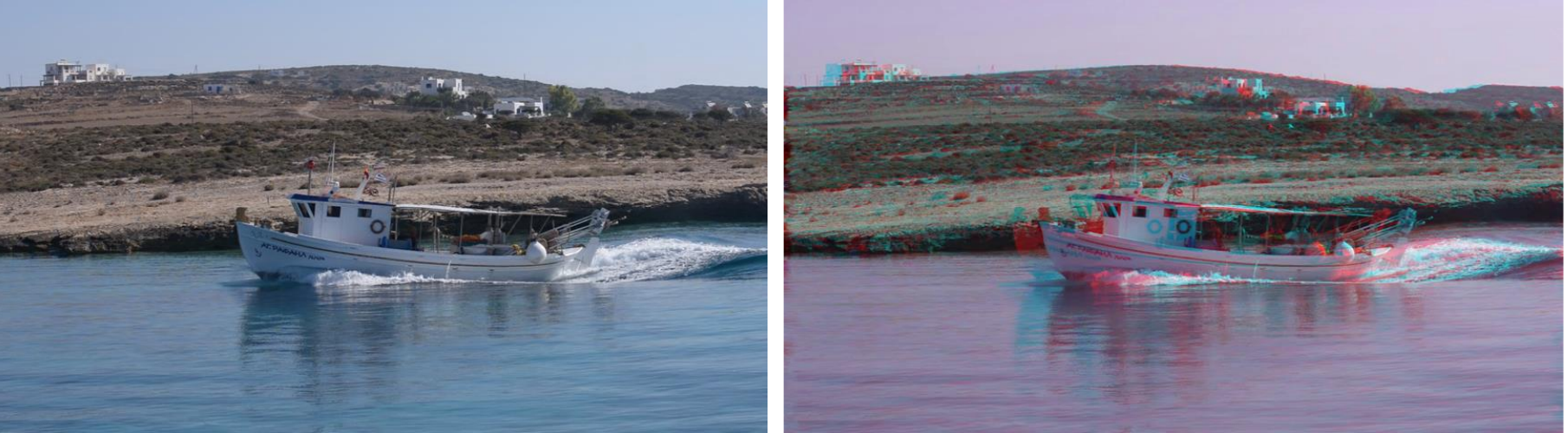}
    \vspace{-6mm}
    \caption{
    \textbf{Stereo view synthesis.}
    Our 3D representation enables novel view synthesis for stereo visualization from monocular video. This demonstrates that Adaptive Gabor primitives in orthographic camera coordinate space capture accurate 3D geometry, enabling immersive applications.
    }
    \label{fig:stereo}
\end{figure}

\paragraph{Stereo View Synthesis.} 
Our explicit representation supports stereo synthesis from monocular input, with improved disparity consistency and plausible geometry, as shown in~\cref{fig:stereo}.

\subsection{Ablation Study}

\input{tables/ablation_gaussian}

\paragraph{Adaptive Gabor Representation.} 
We compare Adaptive Gabor Representation (AGR) to Gaussian and standard Gabor, using the same 1M primitives. As shown in~\cref{tab:ablation_gaussian}, AGR improves high-frequency detail and energy stability, yielding the best PSNR/SSIM/LPIPS among the three configurations.

\input{tables/ablation_spline}

\paragraph{Spline Interpolation.}
We ablate curve interpolation on 50 frames in~\cref{tab:ablation_spline}. B-Spline and Cubic Spline provide some temporal continuity but struggle with nonlinear motion. In contrast, the proposed Cubic Hermite Spline achieves the best performance across metrics, with smoother trajectories and preserved dynamic details.

\begin{figure}[t]
    \centering
    \includegraphics[width=\linewidth]{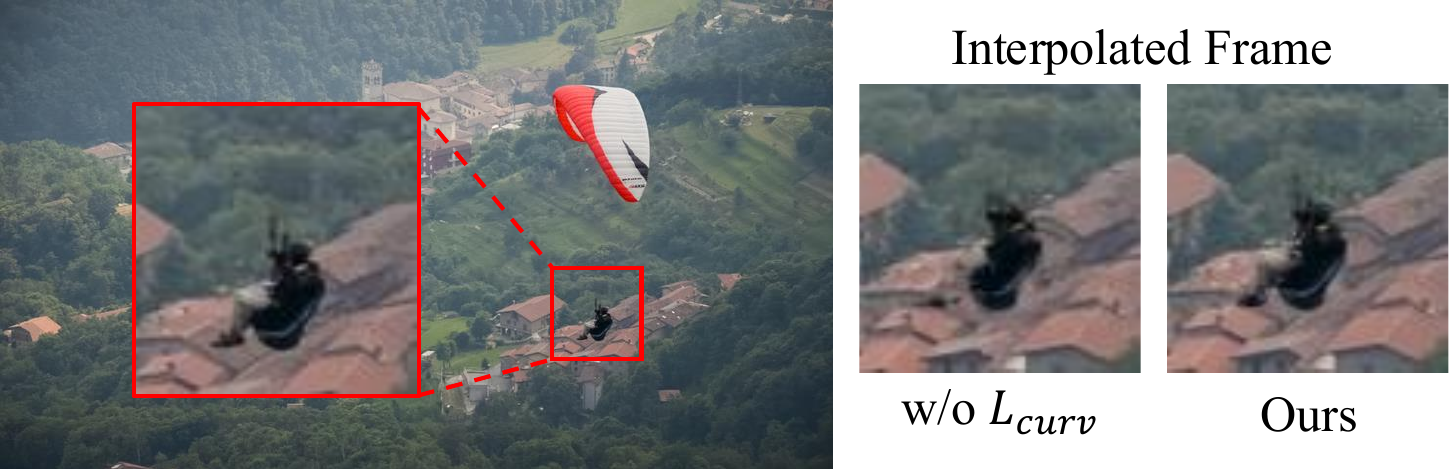}
    \vspace{-6mm}
    \caption{\textbf{Curvature regularization ablation.} 
    \emph{(Left)} Without $\mathcal{L}_\text{curv}$, interpolated frames show motion artifacts from trajectory oscillations. \emph{(Right)} Our method produces smooth, artifact-free interpolation by constraining second-order derivatives, validating the necessity of explicit curvature control for temporal consistency.
    }
    \label{fig:wocurv}
\end{figure}

\paragraph{Curvature Regularization.}
We compare with/without the temporal curvature term \(\mathcal{L}_{\text{curv}}\) in~\cref{fig:wocurv}. Without \(\mathcal{L}_{\text{curv}}\), motion artifacts and tearing appear, while, with \(\mathcal{L}_{\text{curv}}\), interpolation is smoother and more stable, confirming the necessity of explicit curvature control.

\begin{figure}[t]
    \centering
    \includegraphics[width=\linewidth]{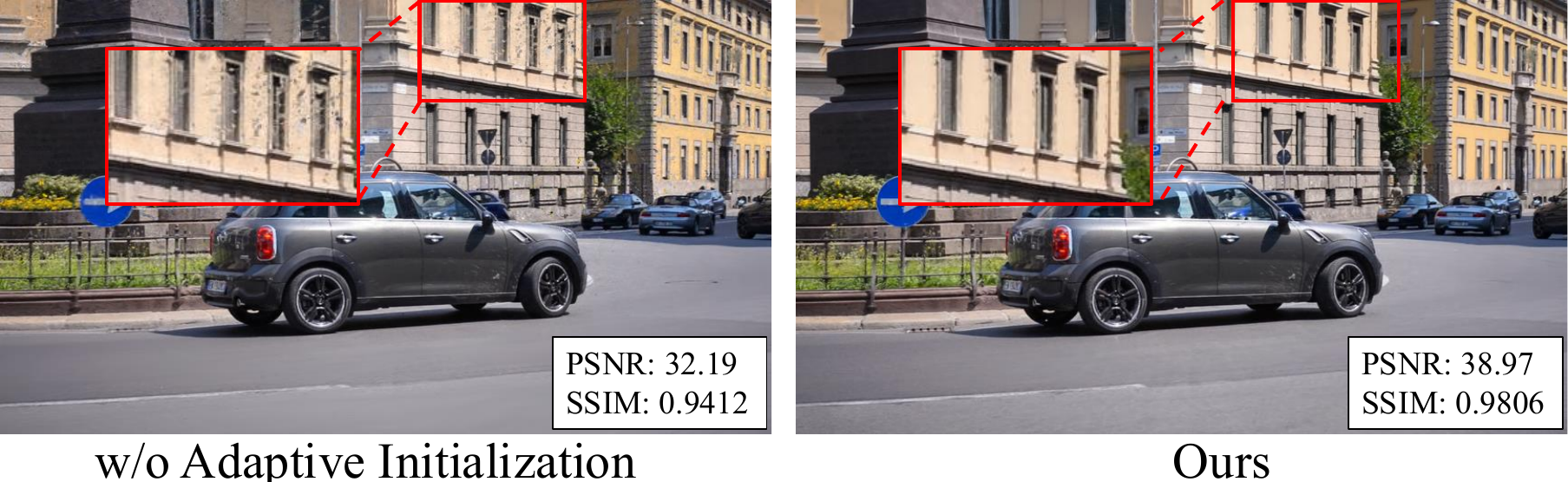}
    \vspace{-6mm}
    \caption{
    \textbf{Adaptive initialization ablation.} (\emph{Left}) Without motion-aware initialization, primitives are poorly distributed, causing blurred details. \emph{(Right)} Our adaptive initialization based on depth, tracking, and masks (\cref{eq:adap_init_1,eq:adap_init_2,eq:adap_init_3}) provides better initial geometry, yielding 6.78 dB improvement and sharp reconstruction.
    }
    \label{fig:adap_init}
\end{figure}

\paragraph{Adaptive Initialization.}
We compare random initialization with our adaptive initialization. As shown in~\cref{fig:adap_init}, the adaptive approach yields denser, temporally coherent initial geometry, reducing flicker and improving early reconstruction quality.

%% file: tables/reconstruction.tex
\begin{table}[t]
\centering
\small
\caption{
\textbf{Quantitative results on Tap-Vid DAVIS~\cite{pont20172017}.}
Our method achieves state-of-the-art performance across all metrics, with 6.86 dB PSNR improvement over the previous best method~\cite{sun2024splatter}, validating our frequency-adaptive primitives with smooth temporal modeling.
}
\label{tab:reconstruction}
    \vspace{-3mm}
\begin{tabular}{l|ccc}
\toprule
{Method} & {PSNR$\uparrow$} & {SSIM$\uparrow$} & {LPIPS↓} \\
\midrule
4DGS~\cite{wu2024deblur4dgs} & 18.12 & 0.5735 & 0.5130 \\
RoDynRF~\cite{liu2023robust} & 24.79 & 0.7230 & 0.3940\\
Deformable Sprites~\cite{ye2022deformable} & 22.83 & 0.6983 & 0.3014 \\
Omnimotion~\cite{wang2023tracking} & 24.11 & 0.7145 & 0.3713\\
CoDeF~\cite{ouyang2024codef} & \cellcolor{yellow!25}26.17 & \cellcolor{yellow!25}0.8160 & \cellcolor{yellow!25}0.2905 \\
Splatter A Video~\cite{sun2024splatter} & \cellcolor{orange!25}28.63 & \cellcolor{orange!25}0.8373 & \cellcolor{orange!25}0.2283\\
Ours & \cellcolor{red!25}35.49 & \cellcolor{red!25}0.9433 & \cellcolor{red!25}0.0723 \\
\bottomrule
\end{tabular}
\end{table}

%% file: tables/ablation_gaussian.tex
\begin{table}[t]
\centering
\small
\caption{
\textbf{Gabor primitive ablation.}
Our Adaptive Gabor with compensation term $b$ outperforms standard Gaussian, naive Gabor variants, validating that energy-aware formulation (\cref{eq:b_equation}) is crucial for stable frequency modeling.
}
\label{tab:ablation_gaussian}
\vspace{-3mm}
\begin{tabular}{l|ccc}
\toprule
{Method} & {PSNR$\uparrow$} & {SSIM$\uparrow$}& {LPIPS$\downarrow$} \\
\midrule
Gaussian & 36.66 & 0.9423 & 0.0421\\
Standard Gabor ($b=0$) & 36.65 & 0.9543 & 0.0345 \\
$1 + S(x)$& 36.50 & 0.9511 & 0.0322 \\
\rowcolor{black!10}
Adaptive Gabor (Ours) & \textbf{37.43} & \textbf{0.9620} & \textbf{0.0242} \\
\bottomrule
\end{tabular}
\end{table}

%% file: tables/ablation_spline.tex
\begin{table}[t]
\centering
\small
\caption{
\textbf{Spline method ablation.}
Our Cubic Hermite Spline with monotone gate outperforms B-Spline and significantly surpasses standard Cubic Spline, which suffers from trajectory oscillations. Explicit velocity control (\cref{eq:ours_spline}) is essential for smooth, artifact-free motion modeling.
}
\label{tab:ablation_spline}
\vspace{-3mm}
\resizebox{\linewidth}{!}{%
\begin{tabular}{lccc}
\toprule
{Methods} & {PSNR$\uparrow$} & {SSIM$\uparrow$}& {LPIPS$\downarrow$} \\
\midrule
B-Spline & 36.68 & 0.9573 & 0.0368\\
Cubic Spline & 32.42 & 0.9073 & 0.0818 \\
\rowcolor{black!10}
Cubic Hermite Spline (Ours) & \textbf{38.98} & \textbf{0.9697} & \textbf{0.0259} \\
\bottomrule
\end{tabular}
}
\end{table}

%% file: 10_conclusion.tex
\section{Conclusion}
\label{sec:conclusion}
We present AdaGaR, a unified framework for temporal continuity and frequency adaptivity in dynamic scene modeling. By extending Gaussian primitives to Adaptive Gabor Representation and employing Cubic Hermite Splines with Temporal Curvature Regularization, our approach captures high-frequency details while ensuring geometric and motion continuity. Experiments demonstrate state-of-the-art performance on Tap-Vid DAVIS with strong generalization across frame interpolation, depth consistency, video editing, and stereo synthesis.

\paragraph{Limitations.}
Despite superior performance, AdaGaR has limitations. The spline-based motion modeling assumes smooth trajectories, potentially causing misalignment under abrupt or highly nonlinear motion. Additionally, Adaptive Gabor Representation may exhibit oscillations in high-frequency regions due to energy constraints. Future work could introduce adaptive temporal control points and motion-aware frequency modulation.

%% file: 12_appendix.tex

\tableofcontents

\section{Activation for Gabor Coefficients}

\subsection{Straight-Through Hard Sigmoid for Frequency Weights}

In Gabor primitives, the frequency coefficients $\omega_i$ must satisfy two requirements: (1) values must be constrained within a learnable range, and (2) gradients must flow back through the activation to enable end-to-end optimization.

To achieve this, we employ a Straight-Through Estimator (STE)~\cite{bengio2013estimating} with a hard sigmoid activation. During the forward pass, we apply hard sigmoid to clip $\omega_i$ into the range $[0, 1]$:
\begin{equation}
\hat{\omega} = \text{clip}\left(\frac{\omega + 1}{2}, 0, 1\right).
\end{equation}

This ensures that the Gabor kernel's frequency modulation remains bounded, preventing unbounded growth that could destabilize energy balance.

However, since the hard clipping operation is non-differentiable, we cannot directly backpropagate through it. Instead, during the backward pass, we use the gradient of the sigmoid function as a surrogate:
\begin{equation}
\frac{\partial L}{\partial \omega} = \frac{\partial L}{\partial \hat{\omega}} \cdot \sigma(\omega)(1 - \sigma(\omega)),
\end{equation}
where $\sigma(\cdot)$ is the standard sigmoid function. This provides a smooth, bounded gradient signal.

The combination of bounded forward pass and smooth backward pass achieves stable training: the forward pass prevents artifacts by constraining frequency weights, while the backward pass enables effective gradient-based optimization. This approach avoids exploding gradients that can arise from unbounded activations.

\section{Proof of Adaptive Degradation to Gaussian}

We prove that our Adaptive Gabor representation naturally degrades to a traditional Gaussian when all frequency weights vanish, demonstrating its adaptive capability between Gaussian and Gabor modes.

\subsection{Mathematical Formulation}

Recall from Eq.~(4) and Eq.~(5) in the main paper, the adaptive modulation function is defined as:
\begin{equation}
S_{\text{adap}}(\mathbf{x}) = b + \frac{1}{N}\sum_{i=1}^{N} \omega_i \cos(f_i \langle \mathbf{d}_i, \mathbf{x} \rangle),
\end{equation}
where the compensation term $b$ is given by:
\begin{equation}
b = \gamma + (1-\gamma)\left(1 - \frac{1}{N}\sum_{i=1}^{N} \omega_i\right),
\end{equation}
with $\gamma \in [0,1]$ as a fixed hyperparameter controlling degradation smoothness, and $1/N$ normalizing the weighted average of multiple waves.

\subsection{Degradation to Gaussian}

Consider the limiting case where all frequency weights approach zero: $\omega_i \to 0$ for all $i \in \{1, \ldots, N\}$.

In this case:
\begin{equation}
\sum_{i=1}^{N} \omega_i \to 0.
\end{equation}

Substituting into the compensation term:
\begin{equation}
b \to \gamma + (1-\gamma)\left(1 - \frac{1}{N} \cdot 0\right) = \gamma + (1-\gamma) \cdot 1 = 1.
\end{equation}

And the modulation term becomes:
\begin{equation}
\frac{1}{N}\sum_{i=1}^{N} \omega_i \cos(f_i \langle \mathbf{d}_i, \mathbf{x} \rangle) \to 0.
\end{equation}

Therefore:
\begin{equation}
S_{\text{adap}}(\mathbf{x}) \to 1 + 0 = 1.
\end{equation}

\subsection{Implication for Opacity}

Since the Gabor-modulated opacity is defined as:
\begin{equation}
\alpha_{\text{Gabor}}(\mathbf{x}) = \mathcal{G}(\mathbf{x}) \cdot S_{\text{adap}}(\mathbf{x}),
\end{equation}
when $S_{\text{adap}}(\mathbf{x}) = 1$, we recover:
\begin{equation}
\alpha_{\text{Gabor}}(\mathbf{x}) = \mathcal{G}(\mathbf{x}) \cdot 1 = \mathcal{G}(\mathbf{x}),
\end{equation}
which is exactly the traditional Gaussian primitive without frequency modulation.

\subsection{Conclusion}

This proof demonstrates that our Adaptive Gabor representation gracefully degrades to a standard Gaussian when frequency content is not needed ($\omega_i \to 0$), while smoothly transitioning to frequency-enhanced Gabor modes when high-frequency details are required ($\omega_i > 0$). This adaptive behavior is crucial for maintaining energy stability across diverse scene regions with varying frequency characteristics.

\begin{figure*}[t]
    \centering
    \includegraphics[width=\linewidth]{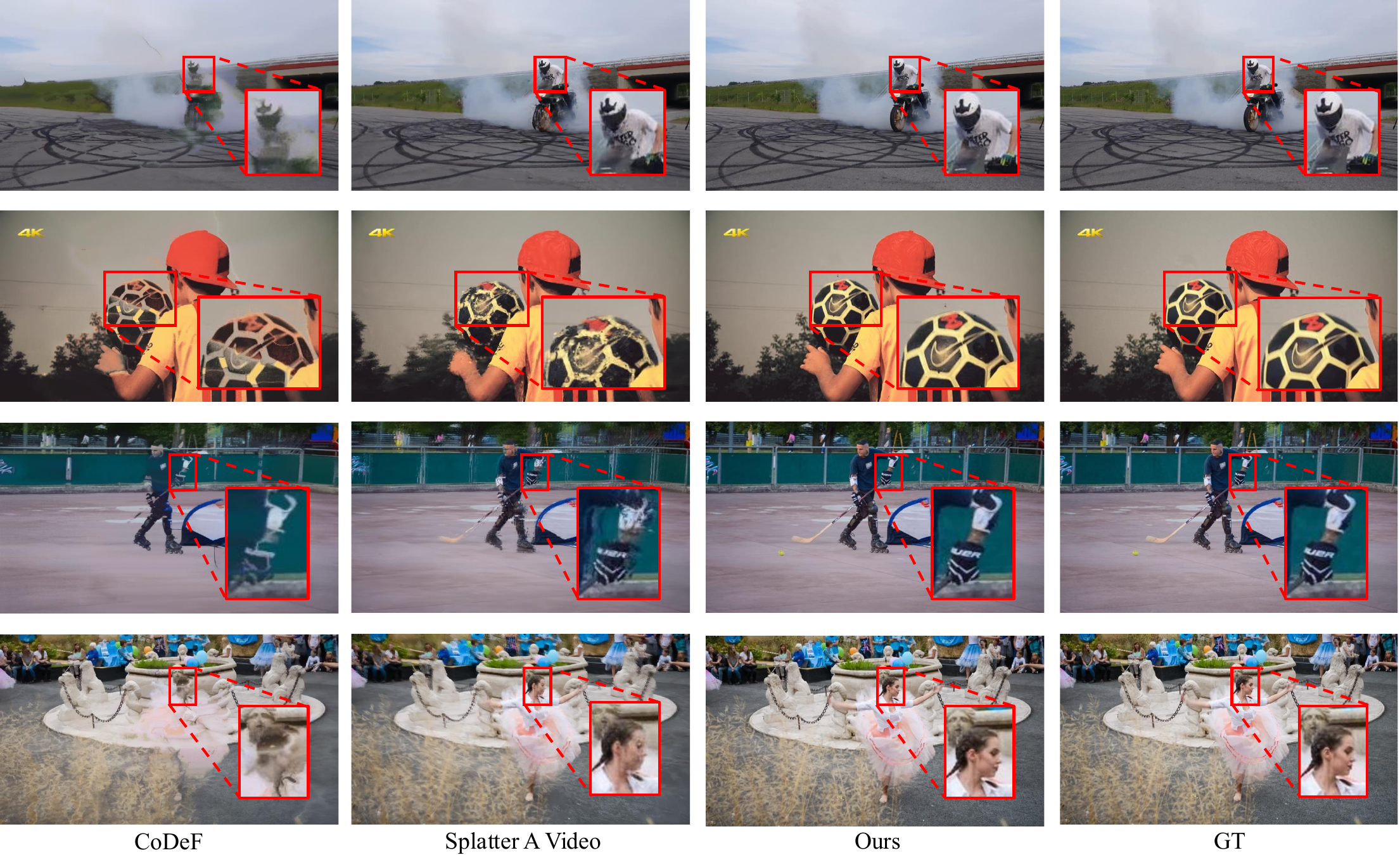}
    \caption{\textbf{Visual comparison on DAVIS dataset.} }
    \label{fig:supple_compare_3}
\end{figure*}

\section{Additional Visual Comparisons and Results}

\begin{figure*}[t]
    \centering
    \includegraphics[width=\linewidth]{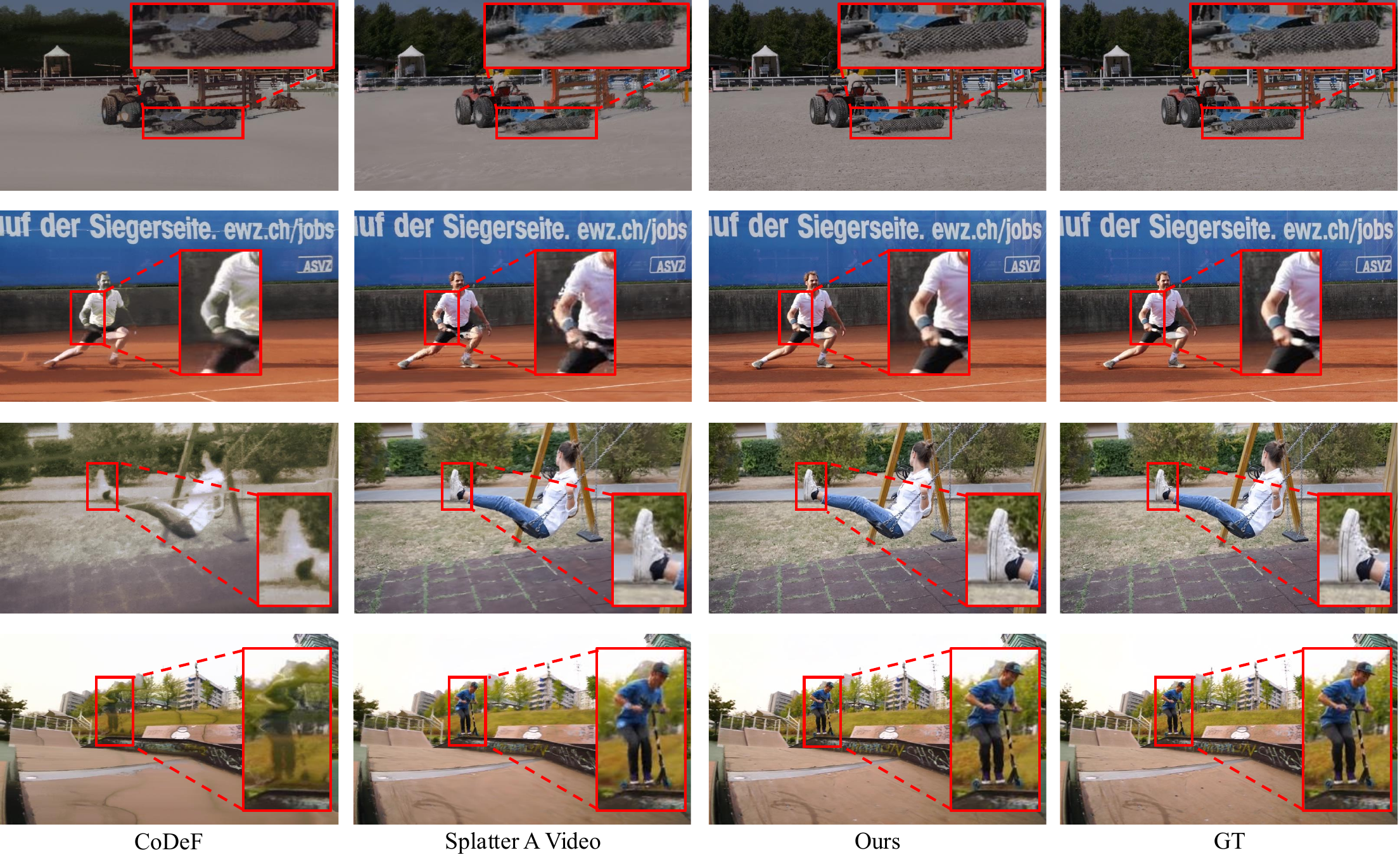}
    \includegraphics[width=\linewidth]{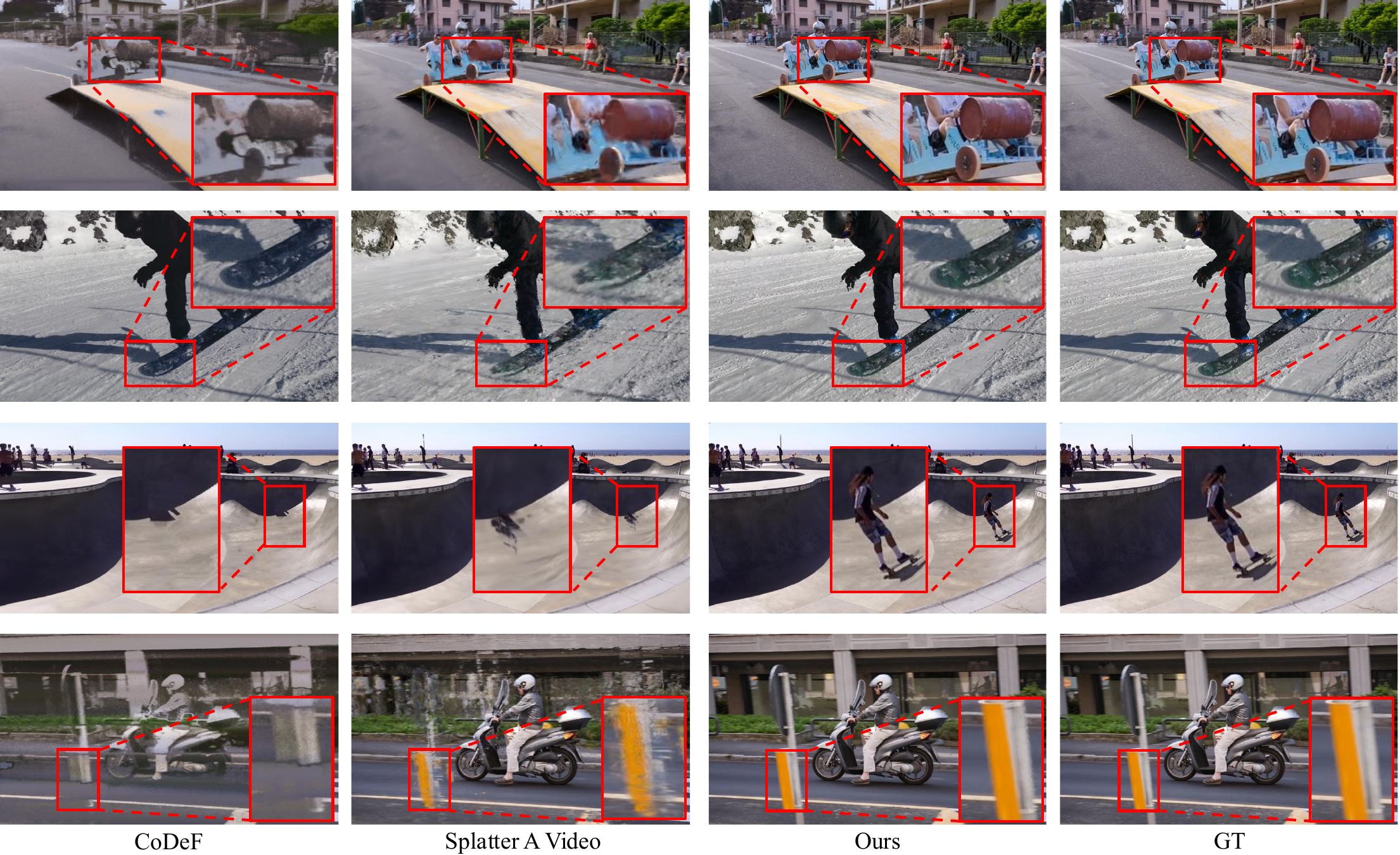}
    \caption{\textbf{Visual comparison on DAVIS dataset.} }
    \label{fig:supple_compare_1}
\end{figure*}

\begin{figure*}[t]
    \centering
    \includegraphics[width=\linewidth]{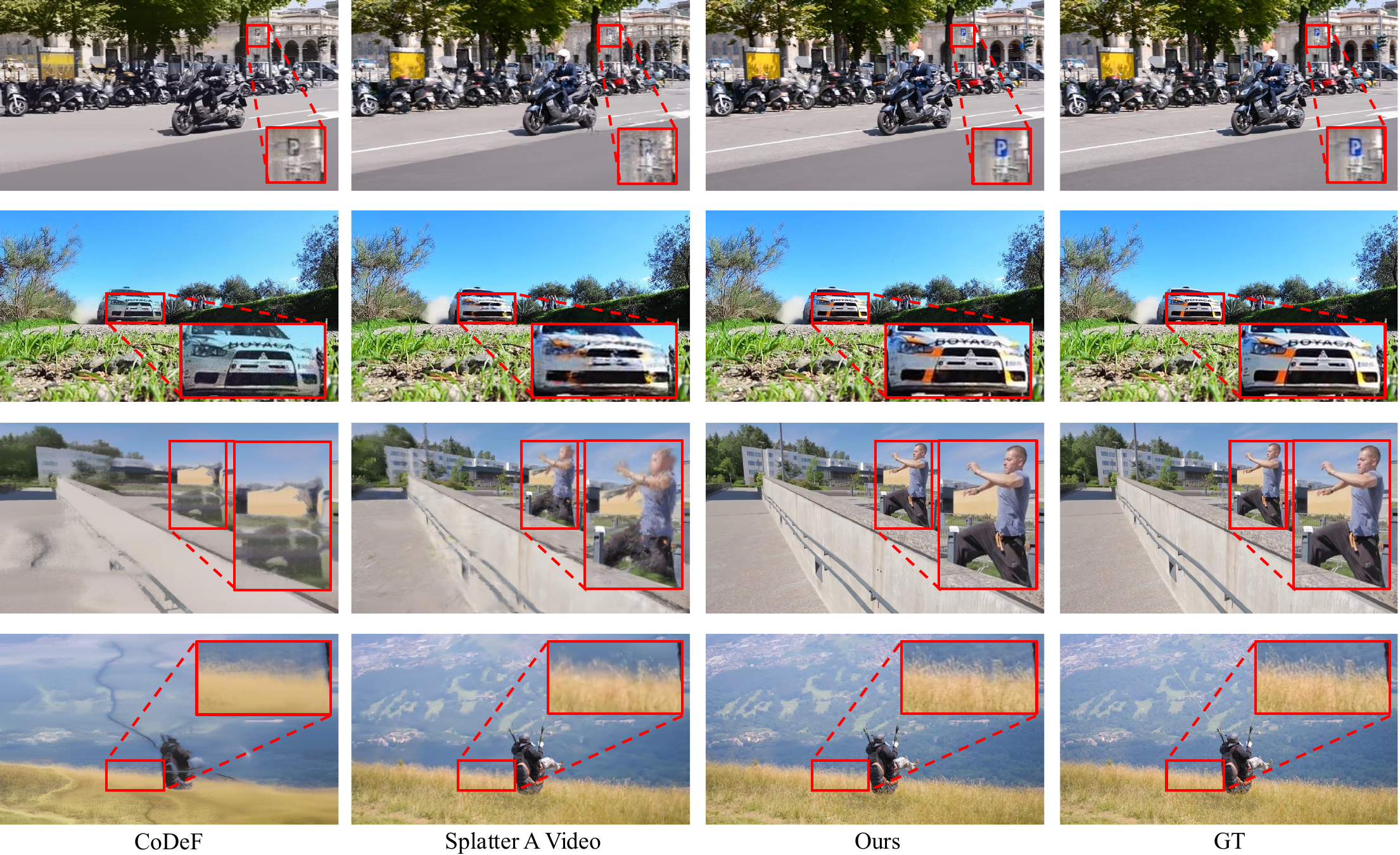}
    \includegraphics[width=\linewidth]{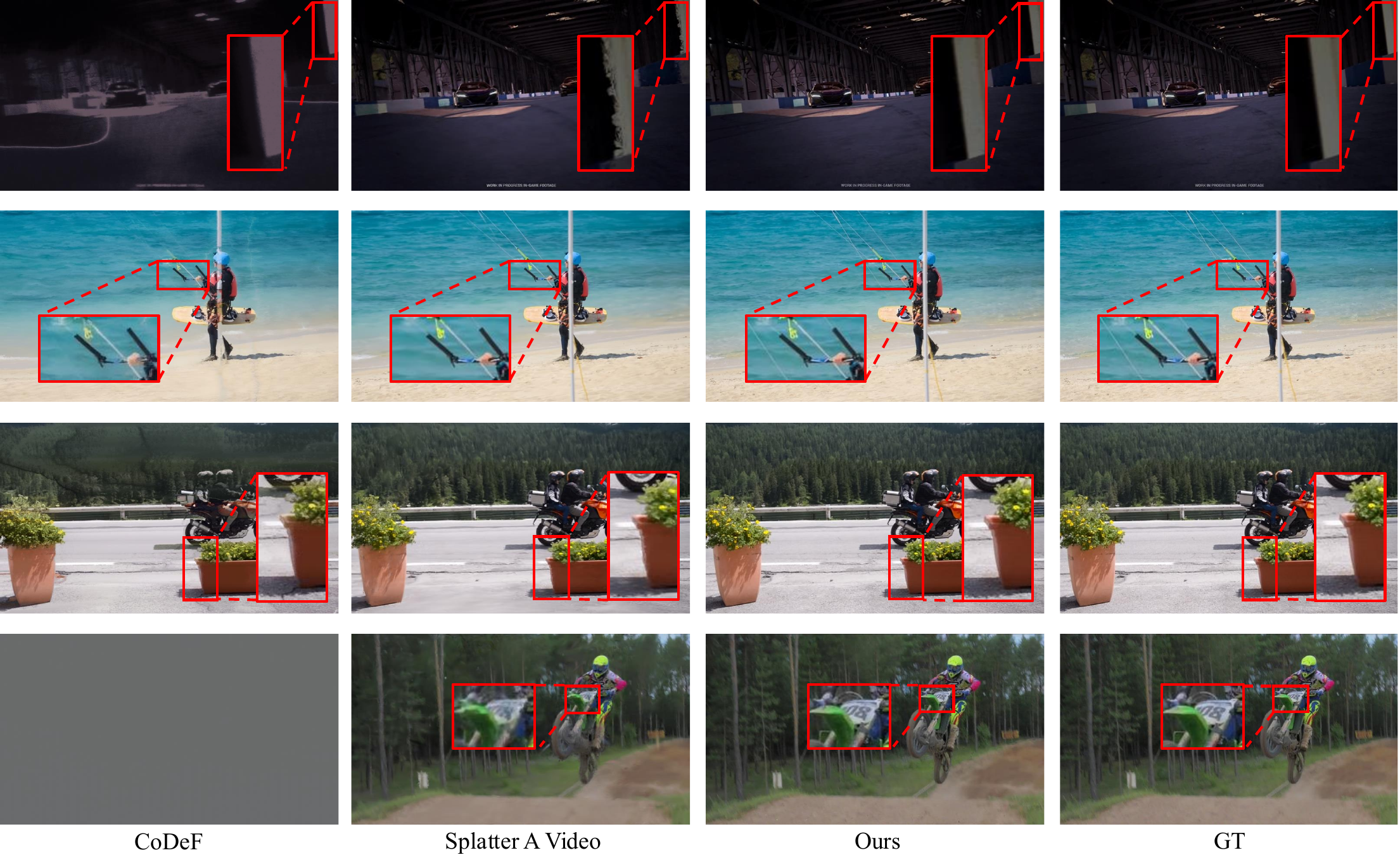}
    \caption{\textbf{Visual comparison on DAVIS dataset.} }
    \label{fig:supple_compare_2}
\end{figure*}

For comprehensive visual comparisons with baseline methods across various dynamic scenes, please refer to \cref{fig:supple_compare_3,fig:supple_compare_1,fig:supple_compare_2}. These figures demonstrate our method's superior performance in preserving high-frequency texture details and maintaining temporal consistency across challenging scenarios including fast motion, occlusions, and complex deformations.

For interactive visualization of downstream application results, including frame interpolation, video editing, and stereo view synthesis, please refer to the supplementary HTML page (\texttt{index.html}). The interactive viewer allows frame-by-frame inspection and video playback to better appreciate the temporal coherence and visual quality of our method.

